\documentclass[conference]{IEEEtran}
\IEEEoverridecommandlockouts
\usepackage{cite}
\usepackage{amsmath,amssymb,amsfonts}
\usepackage{algorithmic}
\usepackage{graphicx}
\usepackage{textcomp}
\usepackage{xcolor}
\usepackage{threeparttable}
\usepackage{booktabs}
\usepackage{tikz}
\usepackage{etoolbox} % 用于调整作者块格式

\makeatletter
\patchcmd{\IEEEauthorblockA}{\@IEEEauthorblockAFont}{\@IEEEauthorblockAFont\small}{}{}
\makeatother

\def\BibTeX{{\rm B\kern-.05em{\sc i\kern-.025em b}\kern-.08em
    T\kern-.1667em\lower.7ex\hbox{E}\kern-.125emX}}
\begin{document}

\title{An Improved Pure Fully Connected Neural Network for Rice Grain Classification}

% \author[2]{Wanke Xia} 
% \author[1*]{Bo Lv}
% \author[1]{Xunwen Xiang}
% \author[2]{Ruoxin Peng} 
% \author[3]{Haoqi Chu}
% \author[2]{Xinlei Zhu}
% \author[2]{Zhiyu Yang}
% \author[2*]{Lili Yang}

% \affil[1]{School of Computer Science and Engineering, Huizhou University, Huizhou, China}
% \affil[2]{College of Information and Electrical Engineering, China Agricultural University, Beijing, China} 
% \affil[3]{College of Land Science and Technology, China Agricultural University, Beijing, China}
% \affil[*]{Corresponding authors: lv\_bobo@qq.com; llyang@cau.edu.cn;}

\author{
\IEEEauthorblockN{Wanke Xia}
\IEEEauthorblockA{\textit{College of Information} \\
\textit{and Electrical Engineering}  \\
\textit{China Agricultural University} \\
Beijing, China \\
2020301010225@cau.edu.cn}

\and

\IEEEauthorblockN{Ruoxin Peng}
\IEEEauthorblockA{\textit{College of Information} \\
\textit{and Electrical Engineering}  \\
\textit{China Agricultural University} \\
Beijing, China \\
2021307150723@cau.edu.cn}
\and
\IEEEauthorblockN{Haoqi Chu}
\IEEEauthorblockA{\textit{College of Land} \\
\textit{Science and Technology}  \\
\textit{China Agricultural University} \\
Beijing, China \\
2020310060310@cau.edu.cn}
\and
\IEEEauthorblockN{Xinlei Zhu}
\IEEEauthorblockA{\textit{College of Information} \\
\textit{and Electrical Engineering}  \\
\textit{China Agricultural University} \\
Beijing, China \\
2021309080108@cau.edu.cn}
\and
\IEEEauthorblockN{Zhiyu Yang}
\IEEEauthorblockA{\textit{College of Information} \\
\textit{and Electrical Engineering}  \\
\textit{China Agricultural University} \\
Beijing, China \\
yangzhiyu@cau.edu.cn}
\and
\IEEEauthorblockN{Lili Yang*}
\IEEEauthorblockA{\textit{College of Information} \\
\textit{and Electrical Engineering}  \\
\textit{China Agricultural University} \\
Beijing, China \\
llyang@cau.edu.cn\\
*Corresponding author}
\and

\IEEEauthorblockN{Bo Lv*}
\IEEEauthorblockA{\textit{School of Computer} \\
\textit{Science and Engineering} \\
\textit{Huizhou University} \\
Huizhou, China \\
lv\_bobo@qq.com\\
*Corresponding author}
\and

% 原来的通讯报销
\IEEEauthorblockN{Xunwen Xiang}
\IEEEauthorblockA{\textit{School of Computer} \\
\textit{Science and Engineering} \\
\textit{Huizhou University} \\
Huizhou, China \\
xiangxw5689@126.com}
}

\maketitle

\begin{abstract}
Rice is a staple food for a significant portion of the world's population, providing essential nutrients and serving as a versatile ingredient in a wide range of culinary traditions. Recently, the use of deep learning has enabled automated classification of rice, improving accuracy and efficiency. However, classical models based on first-stage training may face difficulties in distinguishing between rice varieties with similar external characteristics, thus leading to misclassifications. Considering the transparency and feasibility of model, we selected and gradually improved pure fully connected neural network to achieve classification of rice grain. The dataset we used contains both global and domestic rice images obtained from websites and laboratories respectively. First, the training mode was changed from one-stage training to two-stage training, which significantly contributes to distinguishing two similar types of rice. Secondly, the pre-processing method was changed from random tilting to horizontal or vertical position correction. After those two enhancements, the accuracy of our model increased notably from 97\% to 99\%. In summary, two subtle methods proposed in this study can remarkably enhance the classification ability of deep learning models in terms of the classification of rice grain.
\end{abstract}

\begin{IEEEkeywords}
rice grain, image classification, fully connected neural network, deep learning
\end{IEEEkeywords}

\section{Introduction}
% Rice serves as the staple food and primary source of energy for approximately half of the world's population\cite{ngo2023insights}. With the improvement of people's living standards and the increase in foreign trade, the quality of rice has received considerable attention\cite{aznan2023review}. Sensory evaluation by human assessors is the most widely used method for assessing rice quality. However, it's often influenced by factors such as lighting conditions, visual acuity, and emotions, making it unable to meet the requirements for rapid and objective detection. Traditional methods of rice grain classification often rely on manual labor of sorting, which is quite inefficient and subjective. Also, the identification methods mainly stem from farmers' years of experience, agricultural or agronomy books published by authoritative publishers, experience-sharing posts on the internet, and verbal guidance from technical experts. 
Rice is a staple for half the global population\cite{ngo2023insights}, with its quality gaining attention amid rising living standards and foreign trade\cite{aznan2023review}. Human sensory evaluation, the most common quality assessment method, is flawed by lighting, visual acuity, and emotions, failing rapid, objective detection. Traditional rice classification relies on inefficient, subjective manual sorting, with identification based on farmers' experience, authoritative agricultural books, online posts, and experts' guidance.

%In contrast, deep learning technology can achieve automatic classification by training models, thereby improving the efficiency and accuracy of classification overall. Additionally, through big data analysis and multi-modal fusion technology, deep learning can also help optimize the quality and taste of rice production, becoming an important direction for the future development of rice classification and detection. 

In recent years, classification tasks in the agricultural area based on deep learning techniques involves employing advanced neural networks to automatically categorize rice grain based on their visual features\cite{hasan2023enhancing}. Deep learning models have shown promising results in various image classification tasks, including the precision agriculture field\cite{presti2023current,zhang2024gait}. For example, there has been a formed model of deep convolutional neural network to conduct a binary classification on the superior grades of seeds, with a precision of 93\% and a recall of 95\%\cite{hidayat2023determining}. Meng et al.\cite{meng2023fine} proposed a convolutional neural network classification model based on a self-attention mechanism to differentiate crop species by inputting canopy spectral information, and gained an accuracy of 99.93\%. Patrício et al.\cite{patricio2023oat} applied a dense net architecture for oat grains classification and achieved an accuracy of 99.7\% for oat species identification and an accuracy of 89.7\% for oat varieties classification. Asefa et al.\cite{asefa2023rapid} adopted multivariate data analysis and machine vision for rapid classification of 10 grain varieties, and reached splendid results with 97\% of prediction accuracy and 99\% of precision. Liu et al.\cite{liu2023rice} replaced the feature extraction structure from the SE attention mechanism to a standardized NAM attention mechanism in EfficientNetV2 during the study of discriminating between full grains and grain impurities. This study reduced parameter calculations while maintaining a relatively high accuracy. Tasci et al.\cite{tasci2023efficient} created 8 different quantized neural network (QNN) with the methods of both multi-layer perception (MLP) and Lenet-5 in an attempt to efficiently classify rice varieties, and accomplished an accuracy of 99.87\% with a very small size of memory implemented for the parameters. Chen et al.\cite{chen2024infrared} selected optimal deep neural network (DNN) and convolutional neural network (CNN) after 10,000 iterations to distinguish the origin and variety of rice types. Their study obtained an accuracy of 95.4\% on the calibration set as well as an accuracy of 90.0\% on the validation set when utilizing the optimal DNN model. Thammastitkul et al.\cite{farahnakian2024comparative} implemented DenseNet201 architecture with ImageNet pre-trained weights to improve the grading of Thai Hom Mali rice following standardized grading criteria. Their study achieved an average accuracy of 94.52\% across 6 categories of rice grading. Jeyaraj et al.\cite{jeyaraj2022computer} chose AlexNet architecture for real-time prediction system of rice grading and gained an average accuracy of 98.2\% with 97.6\% sensitivity and 96.4\% specificity. In conclusion, deep learning techniques have a dominant control over the precision agriculture field. 

% However, there are still some obstacles to be improved in various deep learning models applied in the classification of rice grain currently. The main research challenges lie in the insufficient quantity of open-source rice recognition or classification datasets in terms of rice varieties or image numbers, resulting in limited training samples unable to achieve the expected classification effect. In addition to that, fluctuations in practical external factors such as lighting environments during manual image sampling of rice grain, which are not conducive to the feature extraction stage among all the deep learning models. These difficulties have led to less research in the area of classification and grading of rice grains, which is detrimental to the field of seed science and smart agriculture.
However, deep learning models for rice grain classification still have challenges. For example, insufficient open-source datasets (in varieties or image counts) limit training samples, failing to achieve expected results. Additionally, external factors like fluctuating lighting during manual imaging hinder feature extraction. These issues have reduced research on rice classification and grading, harming seed science and smart agriculture.

%The datasets used in this study consists of two parts. The first dataset contains images of 5 categories of rice that are widely grown in the world, and it is accessible through the public kaggle website or the dataset website offered by authors [14]. The second dataset contains images of 6 categories of rice that are widely cultivated in China, and it was collected by our microscope device in the laboratory. 

Therefore, this study proposes a pure fully connected neural network improved with two subtle methods, aiming to address some of the current research challenges mentioned above in the classification of rice grain. The novel aspects in our study could be summarized as follows:

\begin{itemize}
    \item Based on results of the original model, we analyzed the possible reasons why specific rice types are misidentified, thus proposing a multi-stage classification method inspired by the decision tree.
    \item We proposed a pre-processing method for both train-applicable and test-applicable images, which in turn eliminates the need for image enhancement processing.
    \item We collected a first-hand and high-quality image dataset of rice grain in order to tackle the problem of lack of data and verify the effectiveness of proposed enhancements.
\end{itemize}

% part 2
\section{Materials and Methods}

\subsection{Image Acquisition and Description}
% The process of rice grain classification typically involves collecting a large dataset of labeled rice grain images representing different varieties. These images are then used to train the deep learning model to learn the distinguishing features of each rice grain category[15]. There are very few studies on rice classification and recognition, and therefore there are not enough image datasets of different varieties of rice publicly available on the web. 
% The image dataset of this study is mainly derived from a publicly available dataset from previous research and our own dataset collected in the laboratory. The first dataset consists of 5 types of rice, namely Arborio, Basmati, Ipsala, Jasmine, and Karacadag[14]. These types of rice are widely grown and cultivated across the world. The second dataset consists of 6 types of rice, namely Guangdong Simiao rice, Northeastern glutinous rice, Wuchang rice, Panjin crab field rice, Wannian Gong rice, and Yanbian rice. These types of rice are widely grown and cultivated across mainland China. For ease of description, we refer to the first dataset as the global dataset and the second dataset as the domestic dataset.

Our image dataset combines a public dataset from prior research and our lab-collected data. The first (global dataset) includes 5 globally grown rice types: Arborio, Basmati, Ipsala, Jasmine, Karacadag\cite{cinar2022identification}. The second (domestic dataset) has 6 rice varieties widely cultivated in mainland China: Guangdong Simiao, Northeastern glutinous, Wuchang, Panjin crab field, Wannian Gong, Yanbian.

\begin{figure}[htbp]  % [htbp] 控制图片位置（见前文解释）
    \centering  % 图片居中
    \includegraphics[width=0.75\columnwidth]{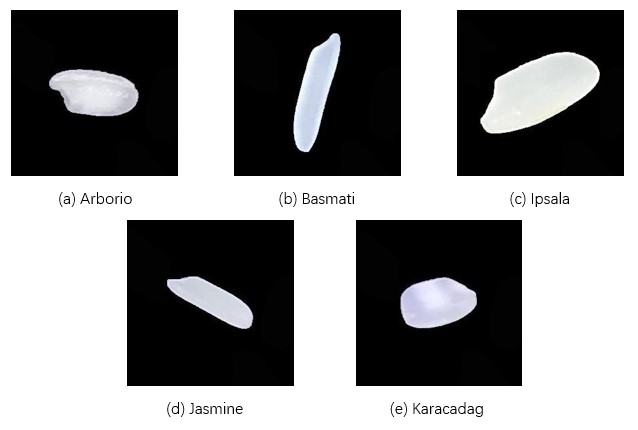}  % 图片文件
    \caption{Sampling images of the global dataset for five types of rice: (a) Arborio, (b) Basmati, (c) Ipsala, (d) Jasmine, (e) Karacadag.}  % 图片标题
    \label{fig:1}  % 标签，用于正文引用
\end{figure}

\begin{table}[htbp]
    \centering
    \caption{Length and width parameters of each type of rice in the global dataset.}
    \label{tab:1}
    \begin{threeparttable}
    \begin{tabular}{lccccc}
        \toprule
        Type & Max Length\tnote{*} & Min Length\tnote{*} & Max Width\tnote{*} & Min Width\tnote{*} \\
        \midrule
        Arborio & 7.5 & 6 & 4 & 3 \\
        Basmati & 11.5 & 8.5 & 4.5 & 3.5 \\
        Ipsala & 11 & 9 & 5.5 & 4 \\
        Jasmine & 10 & 6.5 & 3.5 & 2.5 \\
        Karacadag & 6 & 4.5 & 4 & 3 \\
        \bottomrule
    \end{tabular}
    \begin{tablenotes}
        \footnotesize
        \item[*] Length and width are both in millimeters.
    \end{tablenotes}
    \end{threeparttable}
    \label{tab:rice_parameters}
\end{table}

\begin{figure}[htbp]
    \centering  % 图片居中
    \includegraphics[width=0.75\columnwidth]{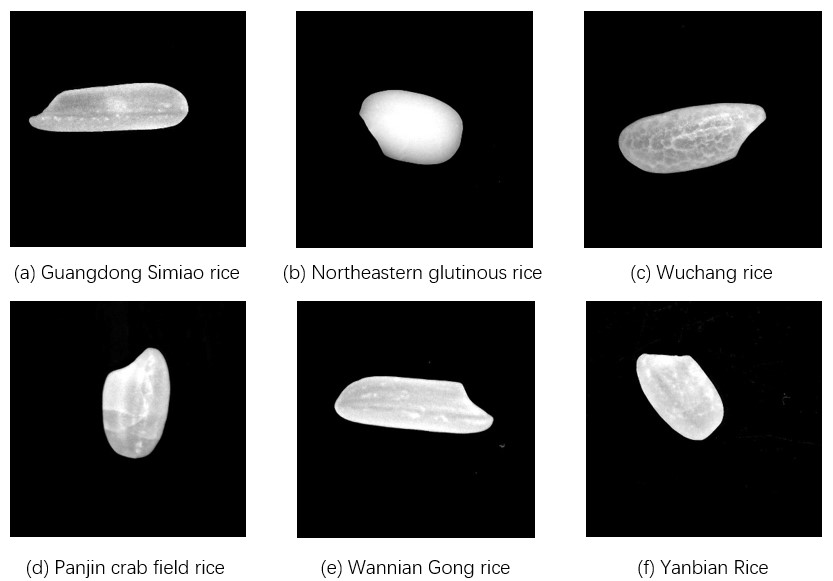}  % 图片文件
    \caption{Sampling images of the domestic dataset for six types of rice: (a) Guangdong Simiao rice, (b) Northeastern glutinous rice, (c) Wuchang rice, (d) Panjin crab field rice, (e) Wannian Gong rice, (f) Yanbian rice.}  % 图片标题
    \label{fig:3}  % 标签，用于正文引用
\end{figure}

\begin{table}[htbp]
    \centering
    \caption{Length and width parameters of each type of rice in the domestic dataset.}
    \begin{threeparttable}
    \begin{tabular}{lcc}
        \toprule
        Type & Average Length\tnote{*} & Average Width\tnote{*} \\
        \midrule
        Guangdong Simiao rice & 6.74 & 1.74 \\
        Northeastern glutinous rice & 4.45 & 2.86 \\
        Wuchang rice & 6.63 & 2.44 \\
        Panjin crab field rice & 4.82 & 2.83 \\
        Wannian Gong rice & 6.81 & 2.20 \\
        Yanbian rice & 4.59 & 2.62 \\
        \bottomrule
    \end{tabular}
    \begin{tablenotes}
        \footnotesize
        \item[*] Length and width are both in millimeters.
    \end{tablenotes}
    \end{threeparttable}
    \label{tab:2}
\end{table}

\subsubsection{The Global Dataset}
In the global dataset, the number of images in each type of rice is 15,000, and the pixels of each image are $250\times250$. For the global dataset, the images are relatively stable in terms of pixels, light, shooting angle and other external disturbances, and the images are all composed of rice in the center with a black non-reflective background. Each type of rice has different morphological characteristics, as shown in Figure \ref{fig:1} and Table \ref{tab:1}. 
% Please note that the public data presented in this study, which refers to the global dataset, can be available at https://www.muratkoklu.com/datasets/.

\subsubsection{The Domestic Dataset}
% Due to the limited types of rice in the public dataset, we first researched rice types that are widely grown in mainland China and purchased the hardware equipment used to capture rice images. The hardware devices for image acquisition in our laboratory are a Sony CMOS sensor which captures images with $1920\times1080$ pixels in real-time, an AOSVI T2-HD228S body microscope with an industrial camera slot on top, a Philips monitor which can be connected to mice, keyboards and industrial cameras via Bluetooth, and other peripherals such as a removable USB flash drive. 
Given the limited rice types in public datasets, we first researched widely grown rice varieties in mainland China and purchased image-capturing hardware. Our lab's equipment includes a Sony CMOS sensor (real-time 1920×1080 pixel imaging), an AOSVI T2-HD228S body microscope with an industrial camera slot, a Philips monitor (Bluetooth-connected to mice, keyboards, and industrial cameras), and peripherals like removable USB drives.

The domestic dataset's image acquisition process includes powering on equipment, fixing a black specimen stage under the objective lens, placeing rice grains flat at its center, adjusting optical parameters for clear magnified images, capturing phenotypic images with an industrial camera, then manually screening and labeling the data.

In the domestic dataset, the number of images in each type of rice is about 3,000, and the pixels of each image are $960\times960$. Raw images needed preprocessing (e.g., cleaning) due to researcher habits, unstable lab lighting, and unfixed platforms. Grayscale and luminance processing was applied to enhance phenotypic features like transparency and texture; processed samples are in Figure \ref{fig:3}.

Additionally, 100 grains per rice type were sampled. Their length and width were measured with vernier calipers and spiral micrometers, averaged to get final data (Table \ref{tab:2}).
% The actual acquisition example process is shown in Figure \ref{fig:2}.

% \begin{figure}[htbp]
%     \centering  % 图片居中
%     \includegraphics[width=0.9\columnwidth]{2.jpg}  % 图片文件
%     \caption{A demonstration of full hardware equipment and actual acquisition operations.}  % 图片标题
%     \label{fig:2}  % 标签，用于正文引用
% \end{figure}

\subsection{Methods of both the basic model and enhancements}
\subsubsection{Fully Connected Neural Network}

\begin{figure}[htbp]
    \centering  % 图片居中
    \includegraphics[width=0.9\columnwidth]{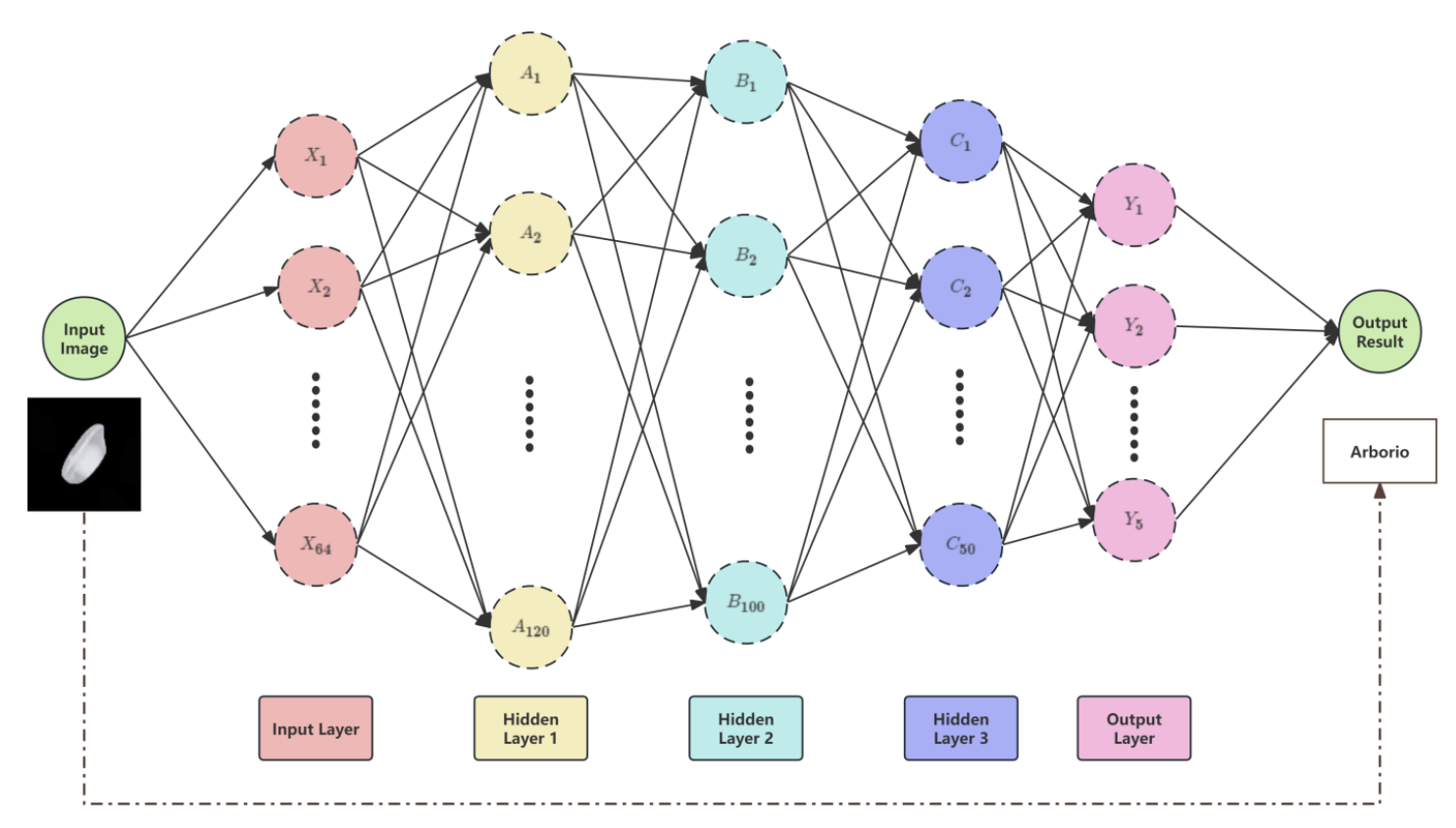}  % 图片文件
    \caption{The general one-stage model architecture using the global dataset as the input images. After the processing of five network layers, the input image will get its classification at the output result.}  % 图片标题
    \label{fig:4}  % 标签，用于正文引用
\end{figure}

Fully connected neural networks have interconnected nodes across layers, with each node in one layer linked to all in the next. Their basic units are neurons, also known as nodes, organized into input, hidden, and output layers.
In this study, the model takes rice images with a black background as input and outputs the closest rice grain classification. For the global dataset, the model has five neural network layers: input layer (64 neurons, fully-connected), hidden layers 1-3 (120, 100, 50 neurons, fully-connected), and output layer (5 neurons, fully-connected, for classification). Its general one-stage architecture is in Figure \ref{fig:4}.

In this model, the activation function of the input layer and the hidden layer adopts the Relu function. The Relu function can be expressed as:
\begin{equation}
    \mathrm{F}(x) = \mathrm{Relu}(x) = \max(0, x)
\end{equation}

In this model, the activation function of the output layer adopts the Softmax function. If let $z$ be the value of output nodes and $C$ be the number of output nodes, the Softmax function can be expressed as:
\begin{equation}
\mathrm{F}(x) = \mathrm{Softmax}(x) = \frac{e^{z_i}}{\sum_{c=1}^{C} e^{z_c}}
\end{equation}

Meanwhile, since the application environment of this model is the multi-classification determination of rice images, the loss function adopts the sparse classification cross entropy. Similarly, we constructed a similar model structure for the domestic dataset.

\subsubsection{Multi-stage Classification Enhancement}
\begin{figure}[htbp]
    \centering  % 图片居中
    \includegraphics[width=0.75\columnwidth]{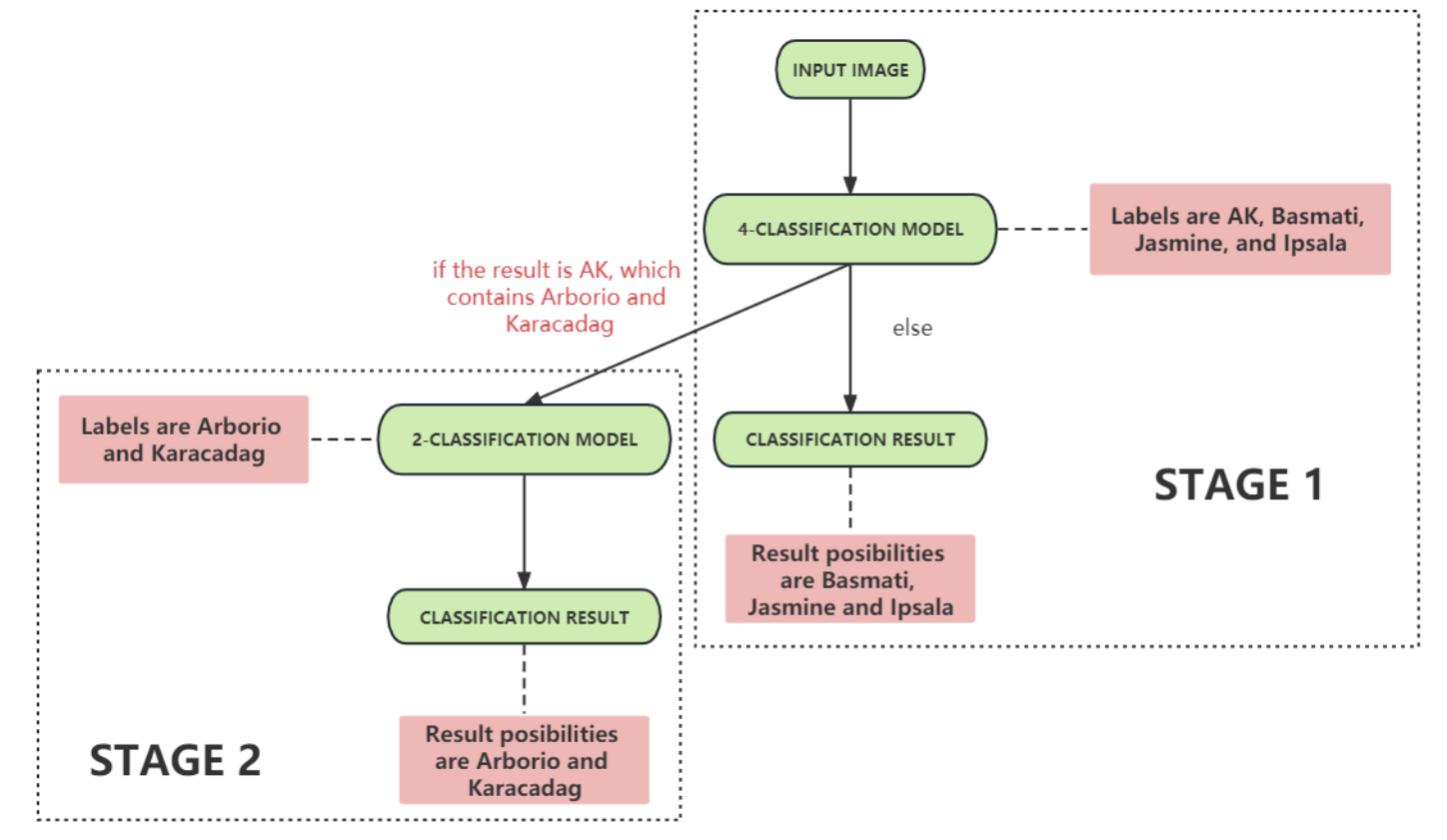}  % 图片文件
    \caption{The overall structure of the model using multi-stage classification enhancement when the global dataset is applied.}  % 图片标题
    \label{fig:5}  % 标签，用于正文引用
\end{figure}

When the two types of rice have similar external characteristics, the one-stage model has a greater possibility of misclassification. So, we propose a multi-stage classification processing method, which is inspired by the decision tree. One of the executable models is shown in Figure \ref{fig:5} and is exemplified by the classification task of the global dataset.

Firstly, the input image is classified by the first fully connected neural network with four possible outputs to get the classification result in the first stage. If the type of this input image may be Arborio or Karacadag, the model will define it as AK, which is the abbreviation of Arborio and Karacadag. So, the possible results in the first stage are AK, Basmati, Jasmine and Ipsala. If the classification result is not AK, the output result will be directly carried out, otherwise it will go to the second stage.

In the second stage, the input image of the first stage will go through the second fully connected neural network with two possible outputs to get the classification results in the second stage. So, the possible results in the first stage are Arborio and Karacadag. It is also worth noting that the four-classification and two-classification models require modification in the number of neurons at the output layer, in order to achieve the ability to normalize the probability of the classification results.

In conclusion, we learned nodes in the intermediate and leaf state from the decision tree, which stands for the mid and final results respectively, and successfully applied these conceptions into fully connected neural networks. When there are more easily misclassified categories present, we can combine them into intermediate state results and later subdivide them into leaf state results.

\subsubsection{Fixed Flipping Enhancement}
\begin{figure}[htbp]
    \centering  % 图片居中
    \includegraphics[width=0.4\columnwidth]{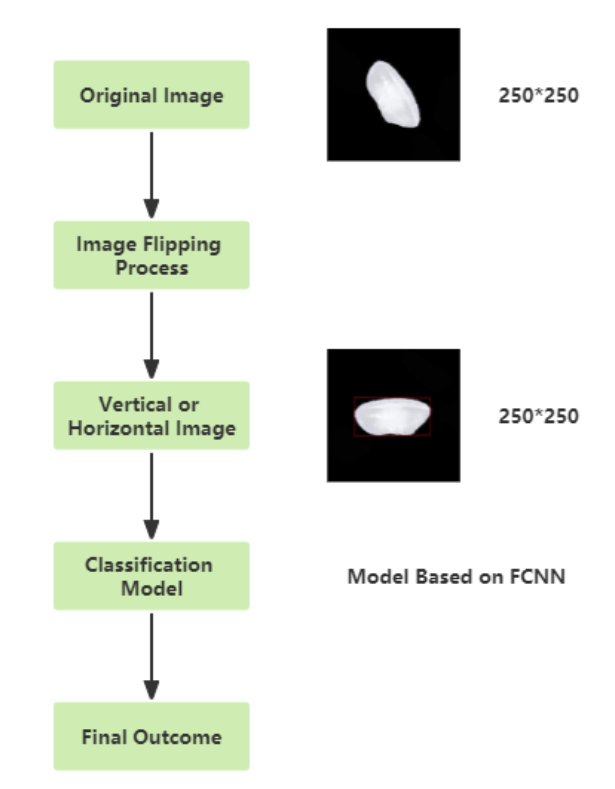}  % 图片文件
    \caption{The schematic diagram of image data input improvement using image rotation to adjust the object to vertical or horizontal location.}  % 图片标题
    \label{fig:6}  % 标签，用于正文引用
\end{figure}

In order to enhance the robustness of the model, classical classification models usually use a large number of images with different placement positions as the training set in order to obtain higher classification accuracy. Especially when the dataset does not have enough images, researchers use data augmentation as a means to expand it, which can be inconvenient to some extent. To be specific, when rice grains with different locations is used as the input images, it may lead to less accurate classification results. To address this issue, this study proposes an input improvement for rotating the image data upfront, as shown in Figure \ref{fig:6}. The original image is first rotated and processed to become an image perpendicular or horizontally parallel to the image boundary. Then, the classification model based on fully connected neural networks (FCNN) completes the training or detection, and finally outputs the classification results.

The principle of image rotation is mainly to utilize the relevant functions to perform the rotation operation and the calibration frame operation on the rice image. First, the rectangular boundary parameters of the rice in the current state are recorded by continuously approximating the boundary of the rice starting from the pixels above, below, left, and right of the rice image. Then, the first step is repeated again and again by rotating and filling the previous image and determining whether the rice rectangular boundary parameter in the current state has reached the state of a very small value. If it has been reached, then output the extreme small value result, otherwise continue the current step. Eventually, it is realized that the position of the rice in the dataset image is transformed from randomly placed to horizontally placed, and the image of the rice, including the red boundary, and the parameters of the boundary calibration box in the horizontally placed state are output.

\section{Results}

\subsection{Setups for Experiments}
The train and valid of aforementioned models are all realized through the Jupyter Notebook platform with kernel of Python 3. The model is based on the Sequential container in Keras framework. Python dependencies are numpy, pandas, matplotlib, os, PIL, sklearn, tensorflow, seaborn, etc. When training, the default value of batch size is 32. The learning rate of the model is automatically adjusted by the optimizers including Adam, SGD, RMSprop, Adadelta, and Nadam.
%The number of iterations is 30. 

\subsection{Performance Metric}
We used the accuracy to evaluate the training results and testing results to determine whether the model is advantageous or not. Accuracy is calculated using the following formula:
\begin{equation}
Accuracy = \frac{TP + TN}{TP + FP + FN + TN}
\end{equation}
where TP refers to the number of samples that are judged to be true and the result is correct, TN refers to the number of samples that are judged to be false and the result is correct, FP refers to the number of samples that are judged to be true and the result is incorrect, and FN refers to the number of samples that are judged to be false and the result is incorrect.

% The accuracy rate is calculated as the ratio between the number of samples correctly predicted by the model and the total number of samples. It provides a simple and intuitive measure of model performance that is easy to understand and interpret. A higher accuracy rate indicates that the model is better able to predict the correct cate-gory in the classification task. To be specific, an accuracy score of 100\% indicates that the model has made all pre-dictions correctly, and an accuracy score of 0\% indicates that the model has made all predictions incorrectly.

\subsection{Performance in the global dataset}
\begin{figure}[htbp]
    \centering  % 图片居中
    \includegraphics[width=0.75\columnwidth]{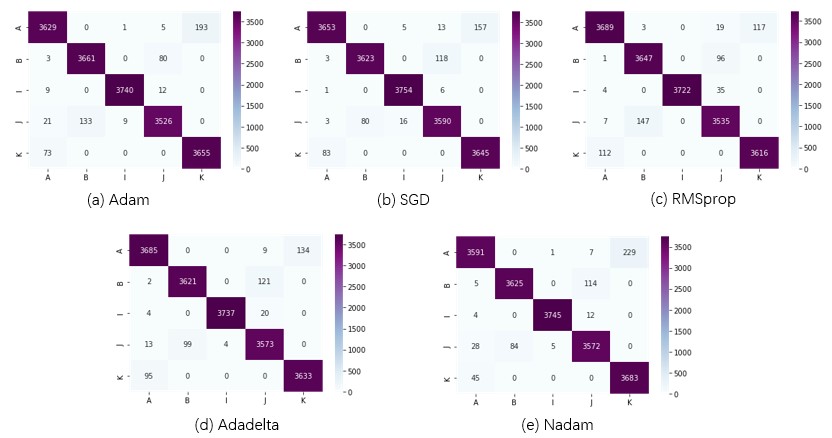}  % 图片文件
    \caption{The confusion matrix of the original model under each optimizer, when the number of iterations is 30.}  % 图片标题
    \label{fig:9}  % 标签，用于正文引用
\end{figure}

We initially used the original model with Adam optimizer and explored the training accuracy and testing accuracy with the maximum number of iterations being 50. Meanwhile, we divided the global dataset into training and testing sets in the ratio of 3:1 and resized the image to $32\times32$. Different optimizers are brought into the model for training and testing. The comparative analysis gives the best performance of SGD optimizer in terms of train accuracy and test accuracy. The results of both the train accuracy and the test accuracy, as well as the confusion matrix of the original model under each optimizer are shown in Table \ref{tab:3} and Figure \ref{fig:9}, respectively.

% The train accuracy and the test ac-curacy are shown in Figure \ref{fig:7}. The train loss and the test loss are shown in Figure \ref{fig:8}.

% \begin{figure}[htbp]
%     \centering  % 图片居中
%     \includegraphics[width=0.8\columnwidth]{86f447233153edd252464bc43077a292.png}  % 图片文件
%     \caption{The Plot of both the train accuracy and the test accuracy of the original model with Adam optimizer, as the number of iterations increases.}  % 图片标题
%     \label{fig:7}  % 标签，用于正文引用
% \end{figure}

% \begin{figure}[htbp]
%     \centering  % 图片居中
%     \includegraphics[width=0.8\columnwidth]{ac0a56b027c6ce713b923ed96e4a5434.png}  % 图片文件
%     \caption{The Plot of both the train loss and the test loss of the original model with Adam optimizer, as the number of iterations increases.}  % 图片标题
%     \label{fig:8}  % 标签，用于正文引用
% \end{figure}

\begin{table}[htbp]
    \centering
    \caption{Accuracy of the original model in the global dataset.}
    \label{tab:3}
    \begin{tabular}{lcc}
        \toprule
        Optimizer & Train Accuracy (\%) & Test Accuracy (\%) \\
        \midrule
        Adam     & 97.62 & 97.13 \\
        SGD      & 98.09 & 97.41 \\
        RMSprop  & 97.38 & 97.11 \\
        Adadelta & 97.67 & 97.33 \\
        Nadam    & 97.61 & 97.15 \\
        \bottomrule
    \end{tabular}
\end{table}

\begin{table}[htbp]
    \centering
    \caption{Accuracy of the model enhanced with both multi-stage classification and fixed flipping on the first stage in the global dataset.}
    \label{tab:4}
    \begin{tabular}{lcc}
        \toprule
        Optimizer & Train Accuracy (\%) & Test Accuracy (\%) \\
        \midrule
        Adam     & 99.11 & 99.18 \\
        SGD      & 99.75 & 99.42 \\
        RMSprop  & 99.59 & 99.27 \\
        Adadelta & 99.66 & 99.37 \\
        Nadam    & 99.71 & 99.48 \\
        \bottomrule
    \end{tabular}
\end{table}

\begin{table}[htbp]
    \centering
    \caption{Accuracy of the model enhanced with both multi-stage classification and fixed flipping on the second stage in the global dataset.}
    \label{tab:5}
    \begin{tabular}{lcc}
        \toprule
        Optimizer & Train Accuracy (\%) & Test Accuracy (\%) \\
        \midrule
        Adam     & 97.55 & 97.25 \\
        SGD      & 98.30 & 97.75 \\
        RMSprop  & 98.35 & 97.40 \\
        Adadelta & 98.37 & 97.40 \\
        Nadam    & 97.91 & 97.50 \\
        \bottomrule
    \end{tabular}
\end{table}

After getting the basic performance, we applied two enhancement methods in the original model. The results of both the first stage and the second stage of the multi-classification model using various types of optimizers, when epoch is set to be 30, are shown in Table \ref{tab:4}. Also, the results of the second stage of the binary classification model using various types of optimizers, when epoch is set to be 30, are shown in Table \ref{tab:5}. We can get that the SGD optimizer performs the best in terms of accuracy in the first and second phase combined. Compared to the results of the original model, the test accuracy has improved from 97.41\% to 99.42\% on the first stage and 97.41\% to 97.75\% on the second stage, when using the SGD optimizer. These results means that two enhancements proposed in this study indeed contribute to the experimental performances to some extent.

\subsection{Performance in the domestic dataset}
Similarly, we processed the data as we did in the global dataset before, such as the dividing ratio of dataset, optimizers, and iterations. After getting and analyzing the basic performance, we finally set epoch to 30 to explore the classification results of the model under different optimizers. The results of both the train accuracy and the test accuracy are shown in Table \ref{tab:6}. The original model performs best when the optimizer is Adam.

\begin{table}[htbp]
    \centering
    \caption{Accuracy of the original model in the domestic dataset.}
    \label{tab:6}
    \begin{tabular}{lcc}
        \toprule
        Optimizer & Train Accuracy (\%) & Test Accuracy (\%) \\
        \midrule
        Adam     & 96.83 & 95.80 \\
        SGD      & 96.53 & 94.58 \\
        RMSprop  & 95.39 & 94.58 \\
        Adadelta & 95.51 & 94.63 \\
        Nadam    & 96.37 & 95.17 \\
        \bottomrule
    \end{tabular}
\end{table}

After getting the basic performance, we applied two enhancement methods in the original model. We noticed that Yanbian rice and Panjin crab field rice have similar phenotypic features, especially opacity and size. So, we further combined them into one type for classification tasks on the first stage, and separated them into two types for classification tasks on the second stage. This operation is similar to the combination and separation of Arborio and Karacadag in the global dataset. The results of the first stage of the multi-classification model using various types of optimizers, when epoch is set to be 30, are shown in Table \ref{tab:7}. Also, the results of the second stage of the binary classification model using various types of optimizers, when epoch is set to be 30, are shown in Table \ref{tab:8}. We can get that the Adam optimizer performs the best in terms of accuracy in the first and second phase combined. Compared to the results of the original model, the test accuracy has improved from 96.83\% to 98.69\% on the first stage and from 96.83\% to 97.34\% on the second stage, when using the Adam optimizer.

\begin{table}[htbp]
    \centering
    \caption{Accuracy of the model enhanced with both multi-stage classification and fixed flipping on the first stage in the domestic dataset.}
    \label{tab:7}
    \begin{tabular}{lcc}
        \toprule
        Optimizer & Train Accuracy (\%) & Test Accuracy (\%) \\
        \midrule
        Adam     & 98.91 & 98.69 \\
        SGD      & 98.95 & 98.30 \\
        RMSprop  & 98.61 & 98.05 \\
        Adadelta & 98.45 & 98.16 \\
        Nadam    & 98.88 & 98.38 \\
        \bottomrule
    \end{tabular}
\end{table}

\begin{table}[htbp]
    \centering
    \caption{Accuracy of the model enhanced with both multi-stage classification and fixed flipping on the second stage in the domestic dataset.}
    \label{tab:8}
    \begin{tabular}{lcc}
        \toprule
        Optimizer & Train Accuracy (\%) & Test Accuracy (\%) \\
        \midrule
        Adam     & 97.62 & 97.13 \\
        SGD      & 98.09 & 97.41 \\
        RMSprop  & 97.38 & 97.11 \\
        Adadelta & 97.67 & 97.33 \\
        Nadam    & 97.61 & 97.15 \\
        \bottomrule
    \end{tabular}
\end{table}

\section{Disccussions}
We mainly discuss the ablation study of both our subtle improvements on the pure fully connected neural network when using the global dataset and the comparison with other classical CNN models.
Similarly, we processed the data as we did in the global dataset before.
%, such as the dividing ratio of dataset, optimizers, and iterations.

\subsection{Only Using the Multi-stage Classification Enhancement}
The results of the first stage of the multi-classification model using various types of optimizers, when epoch is set to be 30, are shown in Table \ref{tab:9}. Also, the results of the second stage of the binary classification model using various types of optimizers, when epoch is set to be 30, are shown in Table \ref{tab:10}.
The experimental results show that when the multi-stage classification method is used in the model trainings stage, the train accuracy and test accuracy of all types of optimizers are improved by 0.1\% to 0.5\%, except RMSprop.

\begin{table}[htbp]
    \centering
    \caption{Accuracy of the model enhanced with only multi-stage classification on the first stage in the domestic dataset.}
    \label{tab:9}
    \begin{tabular}{lcc}
        \toprule
        Optimizer & Train Accuracy (\%) & Test Accuracy (\%) \\
        \midrule
        Adam     & 98.66 & 97.45 \\
        SGD      & 99.31 & 97.37 \\
        RMSprop  & 97.37 & 95.72 \\
        Adadelta & 98.70 & 97.15 \\
        Nadam    & 99.11 & 97.35 \\
        \bottomrule
    \end{tabular}
\end{table}

\begin{table}[htbp]
    \centering
    \caption{Accuracy of the model enhanced with only multi-stage classification on the second stage in the domestic dataset.}
    \label{tab:10}
    \begin{tabular}{lcc}
        \toprule
        Optimizer & Train Accuracy (\%) & Test Accuracy (\%) \\
        \midrule
        Adam     & 97.32 & 96.90 \\
        SGD      & 97.78 & 96.85 \\
        RMSprop  & 98.00 & 97.20 \\
        Adadelta & 98.44 & 97.10 \\
        Nadam    & 98.34 & 96.95 \\
        \bottomrule
    \end{tabular}
\end{table}

\subsection{Only Using the Fixed Flipping Enhancement}
The results of the first stage of the multi-classification model using various types of optimizers, when epoch is set to be 30, are shown in Table \ref{tab:11}. The experimental results show that when the fixed flipping method is used in the image preprocessing stage, the train accuracy and test accuracy of all types of optimizers are improved by 0.5\% to 1.2\%.

\begin{table}[htbp]
    \centering
    \caption{Accuracy of the model enhanced with only fixed flipping in the domestic dataset.}
    \label{tab:11}
    \begin{tabular}{lcc}
        \toprule
        Optimizer & Train Accuracy (\%) & Test Accuracy (\%) \\
        \midrule
        Adam     & 98.25 & 97.72 \\
        SGD      & 99.17 & 98.42 \\
        RMSprop  & 98.72 & 98.32 \\
        Adadelta & 98.90 & 98.28 \\
        Nadam    & 98.65 & 98.28 \\
        \bottomrule
    \end{tabular}
\end{table}

\subsection{Comparison with Other Classical CNN Models}
% Nowadays, pure fully connected neural networks (FCNN) may be not as effective as convolutional neural networks (CNN) in terms of accuracy in the classification tasks. However, they may outweigh CNN models when it comes to the running-time consumption.

\begin{table}[htbp]
  \centering
  \caption{Comparison of both the test accuracy and running-time of our original FCNN model and other classical CNN models.}
  \label{tab:12}
  \begin{threeparttable}
  \begin{tabular}{lcc}
    \toprule
    Model & Test Accuracy (\%) & Running-time \\
    \midrule
    Ours & 97.28 & 37.93 \\
    ResNet50 & 98.99 & 863.03 \\
    GoogLeNet & 99.09 & 936.37 \\
    MobileNet v2 & 99.20 & 691.44 \\
    ConvNeXt-Tiny & 99.11 & 794.05 \\
    \bottomrule
  \end{tabular}
  \begin{tablenotes}
    \item[] \footnotesize * Running-time is in seconds.
  \end{tablenotes}
  \end{threeparttable}
\end{table}

We conducted experiments on the global dataset using a variety of classical CNN models without pre-trained weights in the cuda environment. We set the train iterations as 10 in order to get results faster. Surprisingly, we found that the original FCNN model is trained much faster than those classical CNN models. The results containing both accuracy and running-time can be seen in Table \ref{tab:12}. So, our model may be more timely-efficient when applied in the real-time scenarios.

\section{Conclusion}
% This study mainly builds a model for the classification of rice grain based on fully connected neural networks and discusses the model classification results under multiple optimizers using sparse-categorical-cross-entropy as the loss function. Based on the results of one-stage classification model, we propose multi-stage classification optimization model as well as the input improvement of fixed rotating the image data. In addition, this study collects and labels the images of multiple types of rice widely grown in China mainland. 
We construct a fully connected neural network for rice grain classification and analyze its performance with sparse-categorical-cross-entropy across optimizers. Also, we propose a multi-stage model and fixed image rotation input based on one-stage results, and have collected images of main rice varieties in mainland China.

\section*{Acknowledgement}
This work was supported by the Doctoral Scientific Research Foundation of Huizhou University (No. 2020JB059).

% After training and testing on datasets, good classification results are obtained.

% In summary, the enhancements proposed in this study can be widely applied in the classification of precision agriculture field, especially in the rice grain. The model has a large degree of prospect in improving the quality and added value of rice, helping farmers to plant scientifically, promoting the circulation and marketization of agricultural products, and promoting the optimization of the structure of agricultural industry. 

\bibliographystyle{IEEEtran}
\bibliography{referencesflie}

\end{document}